\documentclass{article}
\usepackage[final]{nips_2016}

\usepackage[utf8]{inputenc} % allow utf-8 input
\usepackage[T1]{fontenc}    % use 8-bit T1 fonts
\usepackage{hyperref}       % hyperlinks
\usepackage{url}            % simple URL typesetting
\usepackage{booktabs}       % professional-quality tables
\usepackage{amsfonts}       % blackboard math symbols
\usepackage{microtype}      % microtypography
\usepackage{nicefrac}      % compact symbols for 1/2, etc.
\usepackage{authblk}

\usepackage{amsmath,amssymb}
\newcommand{\ts}{\textstyle}

\usepackage{natbib}
\setcitestyle{authoryear,open={(},close={)}}

\usepackage{graphicx}
\graphicspath{{figures/}}

\usepackage[font=small]{caption}
\usepackage{float}
\usepackage{algorithm}
\usepackage[noend]{algpseudocode}
\algrenewcommand\algorithmicfunction{\textbf{def}}
\algrenewcommand\algorithmicrequire{\textbf{Input:}}
\algrenewcommand\algorithmicensure{\textbf{Output:}}

\usepackage{color}

\title{Supervised topic models for clinical interpretability}
\author[1]{\textbf{Michael C. Hughes}}
\author[1]{\textbf{Huseyin Melih Elibol}}
\author[2]{\textbf{Thomas McCoy, M.D.}}
\author[2]{\textbf{Roy Perlis, M.D.}}
\author[1]{\textbf{Finale Doshi-Velez}}
\affil[1]{School of Engineering and Applied Sciences, Harvard University, Cambridge, MA, USA}
\affil[2]{Massachusetts General Hospital, Boston, MA, USA}

\begin{document} 
\maketitle
% Reset vertical space for equations 
% (must be after \begin{document})
\setlength{\abovedisplayskip}{2pt plus 3pt}
\setlength{\belowdisplayskip}{2pt plus 3pt}

\begin{abstract}
Supervised topic models can help clinical researchers find interpretable cooccurence patterns in count data that are relevant for diagnostics.
However, standard formulations of supervised Latent Dirichlet Allocation have two problems.  First,
when documents have many more words than labels, the influence of the labels will be negligible.  Second, due to conditional independence assumptions in the graphical model the impact of supervised labels on the learned topic-word probabilities is often minimal, leading to poor predictions on heldout data.
We investigate penalized optimization methods for training sLDA that produce interpretable topic-word parameters and useful heldout predictions, using recognition networks to speed-up inference. We report preliminary results on synthetic data and on predicting successful anti-depressant medication given a patient's diagnostic history.
\end{abstract}

% Sec. 1
\section{Introduction}

Abundant count data---procedures, diagnoses, meds---are produced during clinical care. An important question is how such
data can assist treatment decisions.  Standard pipelines usually involve some dimensionality reduction---there are over
14,000 diagnostic ICD9-CM codes alone---followed by training
on the task of interest.  Topic models such as latent Dirichlet allocation (LDA) \citep{blei2012topicmodels} are
a popular tool for such dimensionality reduction
(e.g. \citet{paul2014discovering} or \citet{ghassemi2014unfolding}).
However, especially given noise and irrelevant signal in the data,
this two-stage procedure may not produce the best predictions; thus
many efforts have tried to incorporate observed labels into the dimensionality reduction model. The most natural extension is \emph{supervised LDA} \citep{blei2008sLDA}, though other attempts exist~\citep{zhu2012medlda,
  lacoste2009disclda}.

Unfortunately, a recent survey by \citet{halpern2012comparison} finds
that many of these approaches have little benefit, if any, over standard LDA.
We take inspiration from
recent work \citep{chen2015bplda} to develop an optimization algorithm that prioritizes document-topic embedding functions useful for heldout data and allows a penalized balance of generative
and discriminative terms, overcoming problems with traditional maximum likelihood point estimation or more Bayesian approximate posterior estimation.  We extend this work with recognition network that allows us to scale to a data set of over 800,000 patient encounters via an approximation to the ideal but expensive embedding required at each document.
\section{Methods}
We consider models for collections of $D$ documents, each drawn from
the same finite vocabulary of $V$ possible word types. Each document
consists of a supervised binary label $y_d \in \{0, 1\}$ (extensions to non-binary labels are straightforward) and $N_d$
observed word tokens $x_d = \{ x_{dn} \}_{n=1}^{N_d}$, with each word
token an indicator of a vocabulary type. We can compactly write $x_d$
as a sparse count histogram, where $x_{dv}$ indicates the \emph{count}
of how many words of type $v$ appear in document $d$. 

\subsection{Supervised LDA and Its Drawbacks} 
Supervised LDA \citep{blei2008sLDA} is a generative model with the following log-likelihoods:
\begin{align}
\log p(x_d | \phi, \pi_d) &=
\log \mbox{Mult}( x_d | N_d, \ts \sum_{k=1}^K \pi_{dk} \phi_k ) 
= \sum_{v=1}^V x_{dv} \log \left( \ts \sum_{k=1}^K \pi_{dk} \phi_{kv} \right)
\\ \notag
\log p(y_d | \pi_d, \eta) &= \log \mbox{Bern}(y_d | \sigma(\eta^T \pi_d) ) = 
y_d \log \sigma(\eta^T\pi_d) 
+ (1-y_d) \log (1 - \sigma(\eta^T \pi_d) )
\end{align}
where $\pi_{dk}$ is the probability of topic $k$ in document $d$, $\phi_{kv}$ is the probability of word $v$ in topic $k$, $\eta_k$ are coefficients for predicting label $y_d$ from doc-topic probabilities $\pi_d$ via logistic regression, and $\sigma(\cdot)$ is the sigmoid function. Conjugate Dirichlet priors $p(\pi_d)$ and $p(\phi_k)$ can be easily incorporated.

For many applications, we wish to either make predictions of $y_d$ or inspect the topic-word probabilities $\phi$ directly. In these cases, point estimation is a
simple and effective training goal, via the objective:
\begin{align}
\max_{\phi, \pi, \eta}~
w_y \Big( \sum_{d=1}^D \log p(y_d | \eta, \pi_d) \Big)
+ w_x \Big(\log p(\phi) + \sum_{d=1}^D \log p(x_d | \pi_d, \phi) + \log p(\pi_d) \Big)
\label{eq:def_penalized_map_objective}
\end{align}
We include penalty weights $w_x > 0, w_y >
0$ to allow adjusting the importance of the unsupervised data term and the
supervised label term. \citet{taddy2012topicmodelmapestimation} gives a coordinate ascent algorithm for the totally unsupervised objective ($w_x=1, w_y=0$), using natural parameterization to obtain simple updates. Similar algorithms exist for all valid penalty weights. 

\begin{figure}
{\scriptsize
\begin{tabular}{r c r c}
\begin{minipage}{1.1cm}
true topics:
\end{minipage}
&
\begin{minipage}{5cm}
\includegraphics[width=\textwidth]
{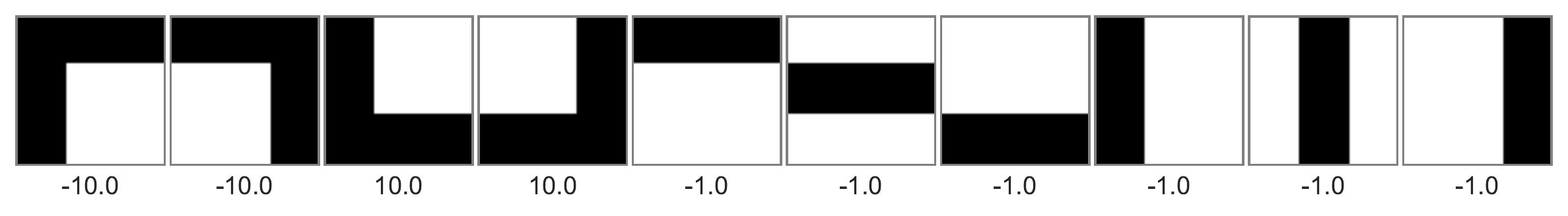}
\end{minipage}
&
\begin{minipage}{1.6cm}
~~~~example docs:
\end{minipage}
&
\begin{minipage}{5cm}
\includegraphics[width=\textwidth]
{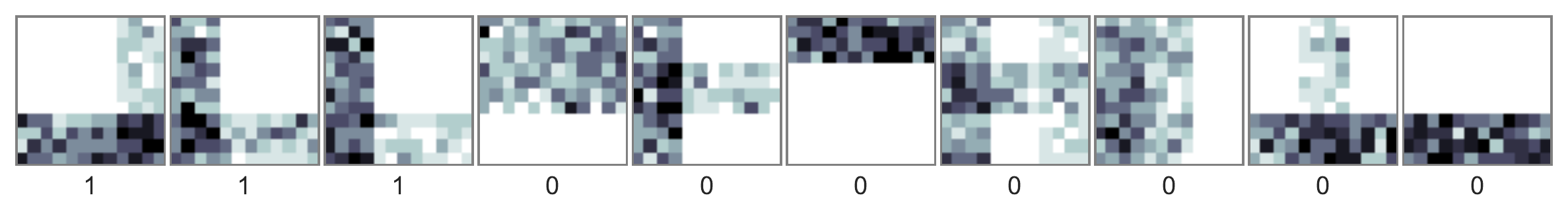}
\end{minipage}
\end{tabular}
\\
\begin{tabular}{p{1cm} c c c}
& Train with Instantiated $\pi$
& Train with Ideal Embedding
& Train with Approx. Embedding
\\
& $\max \log p(y, x | \eta, \phi, \pi)$
& $\max \log p(y, x | \eta, \phi, f^*(x, \phi))$
& $\max \log p(y, x | \eta, \phi, f^{\lambda}(x, \phi))$ 
\\

\begin{minipage}{1cm}
\begin{align*}
w_x = 1 \\
w_y = 0
\end{align*}
\end{minipage}
&
\begin{minipage}{4cm}
\includegraphics[width=\textwidth]{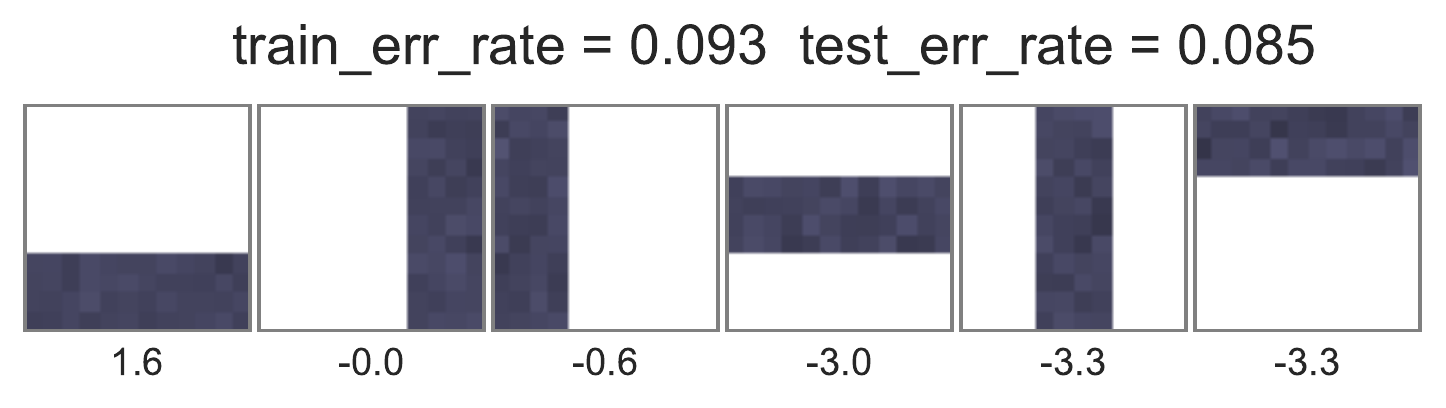}
\end{minipage}
&
\begin{minipage}{4cm}
\includegraphics[width=\textwidth]{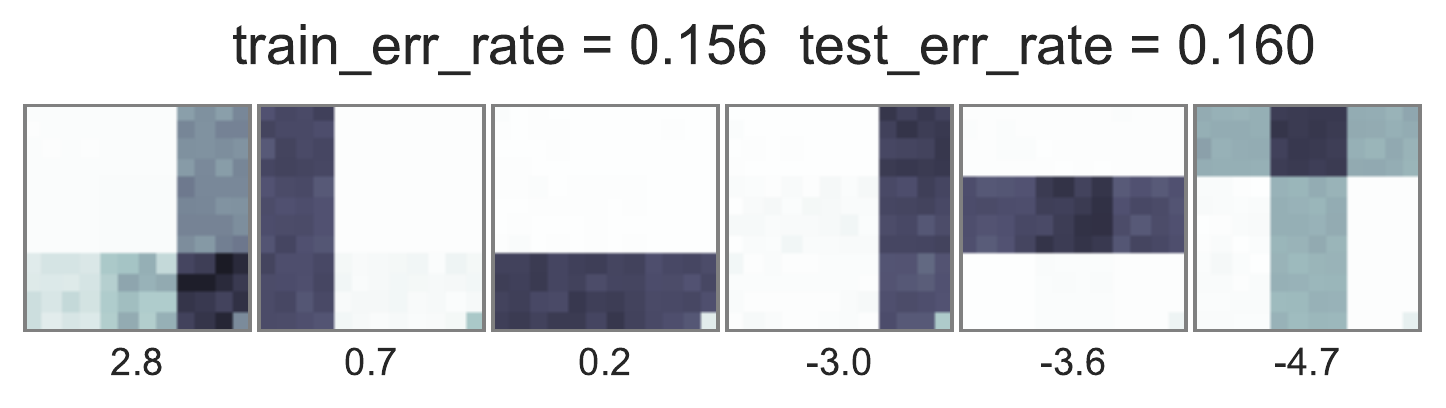}
\end{minipage}
&
\begin{minipage}{4cm}
\includegraphics[width=\textwidth]{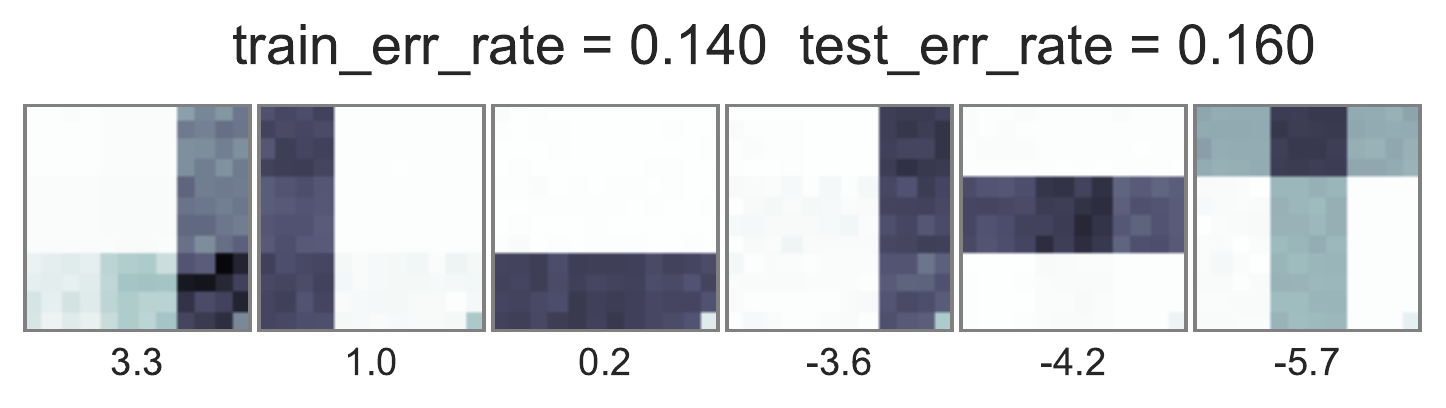}
\end{minipage}
\\
%%%%%%%%%%%%%%%%%%%%%

\begin{minipage}{1cm}
\begin{align*}
w_x = 1 \\
w_y = 1
\end{align*}
\end{minipage}
&
\begin{minipage}{4cm}
\includegraphics[width=\textwidth]{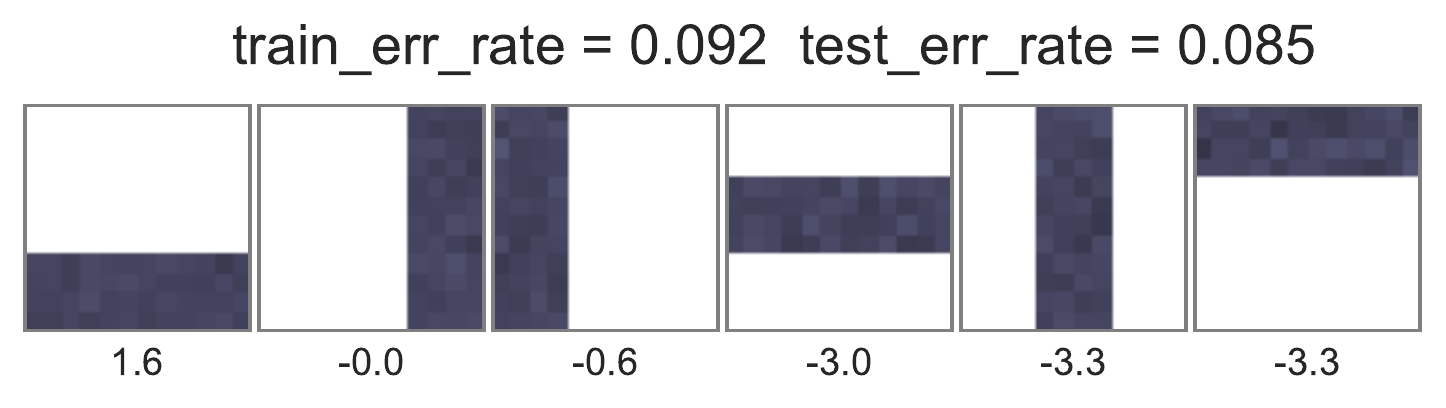}
\end{minipage}
&
\begin{minipage}{4cm}
\includegraphics[width=\textwidth]{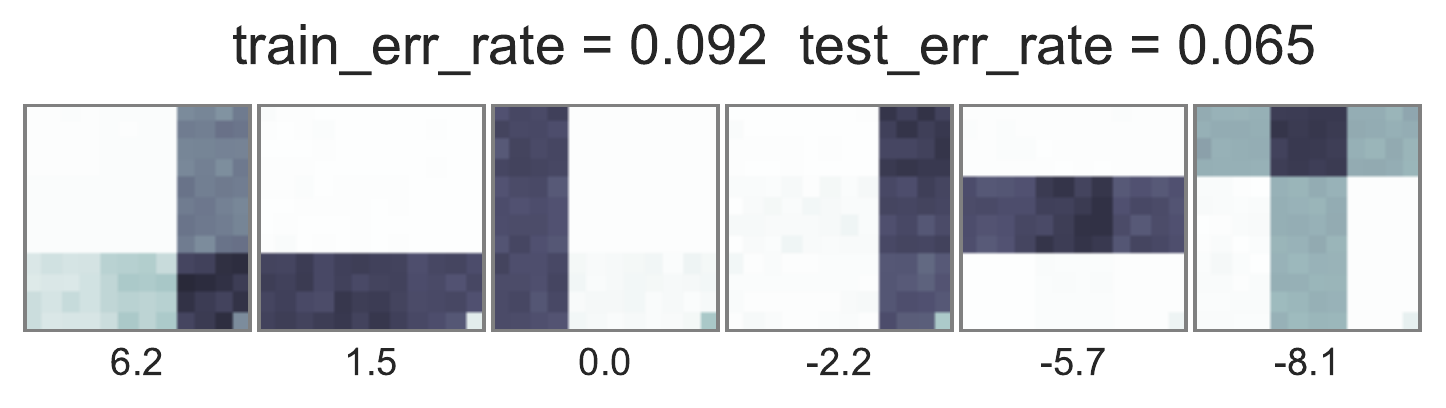}
\end{minipage}
&
\begin{minipage}{4cm}
\includegraphics[width=\textwidth]{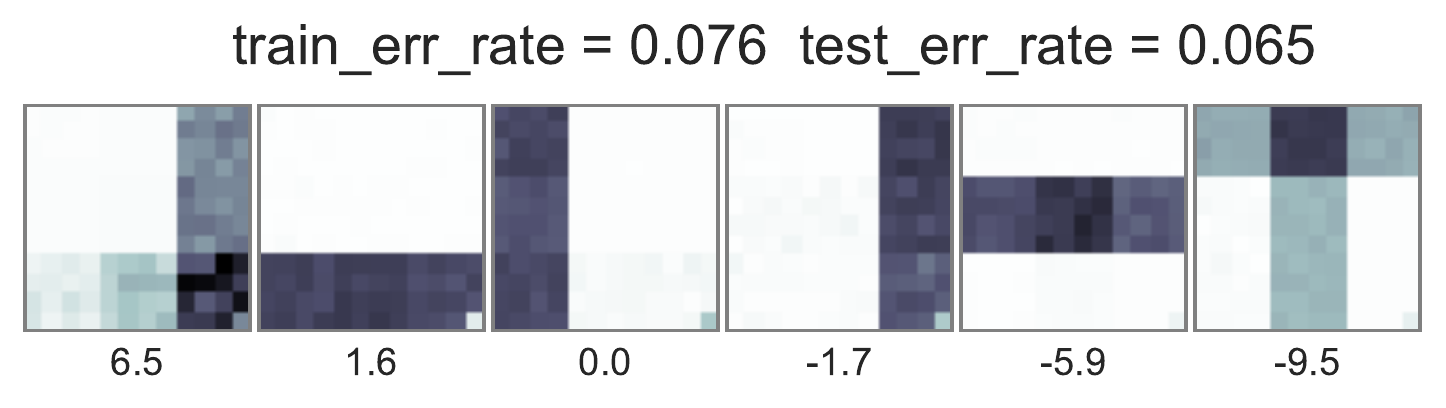}
\end{minipage}
\\

%%%%%%%%%%%%%%%%%%%%%
\begin{minipage}{1cm}
\begin{align*}
w_x = .01 \\
w_y = ~~~1
\end{align*}
\end{minipage}
&
\begin{minipage}{4cm}
\includegraphics[width=\textwidth]{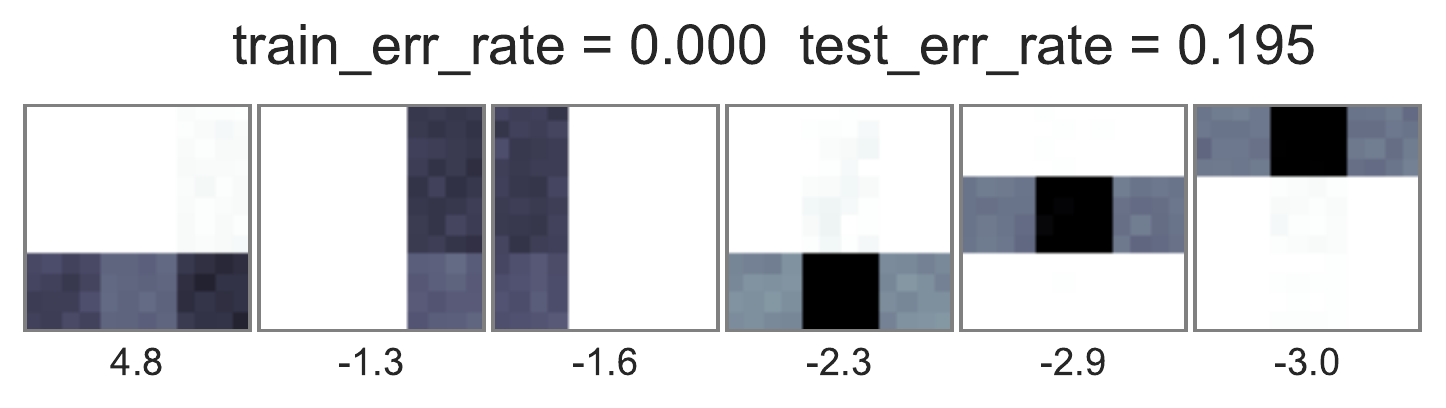}
\end{minipage}
&
\begin{minipage}{4cm}
\includegraphics[width=\textwidth]{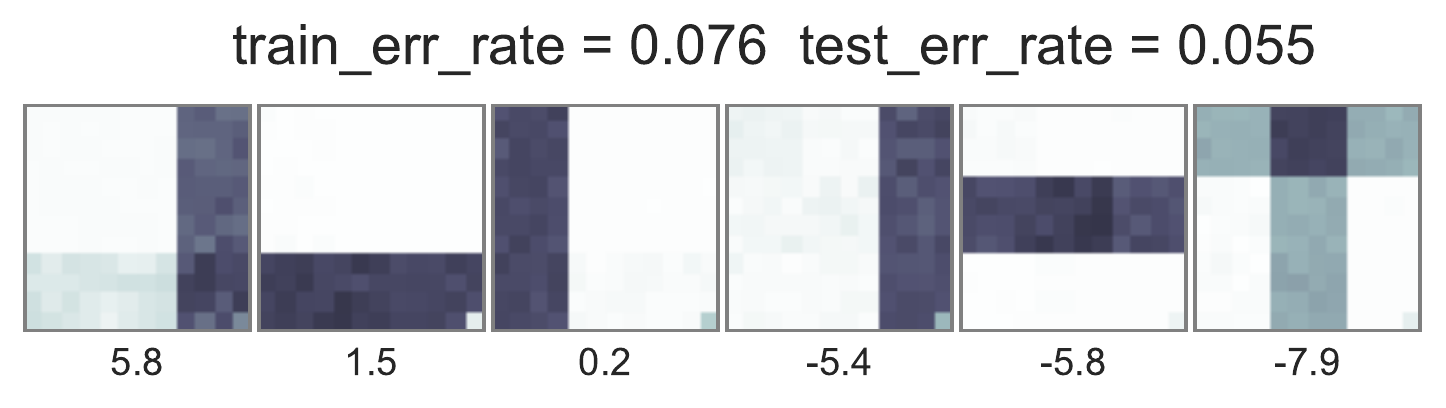}
\end{minipage}
&
\begin{minipage}{4cm}
\includegraphics[width=\textwidth]{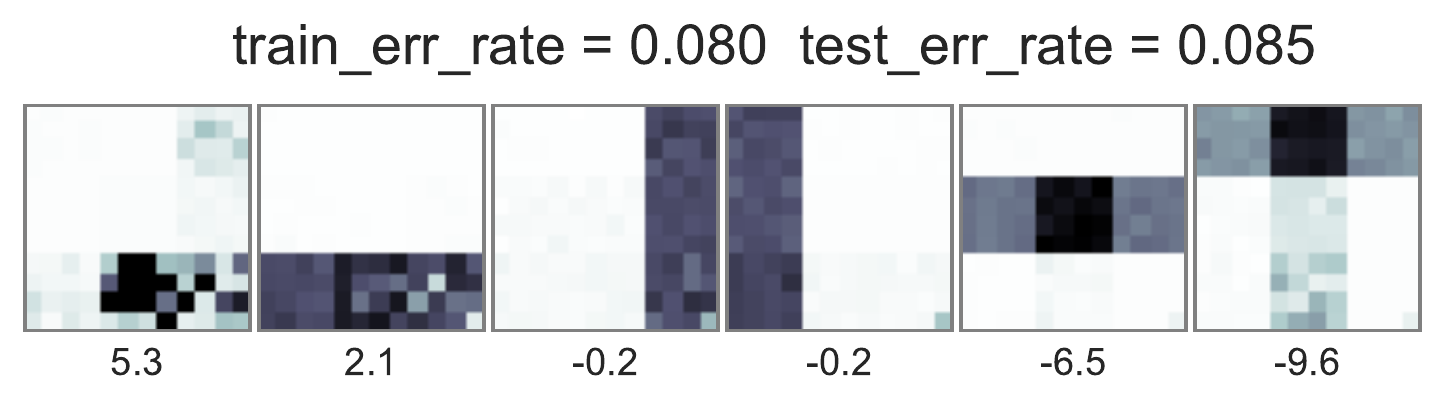}
\end{minipage}
\\

%%%%%%%%%%%%%%%%%%%%%
\begin{minipage}{1cm}
\begin{align*}
w_x = 0 \\
w_y = 1
\end{align*}
\end{minipage}
&
\begin{minipage}{1cm}
N/A
\end{minipage}
&
\begin{minipage}{4cm}
\includegraphics[width=\textwidth]{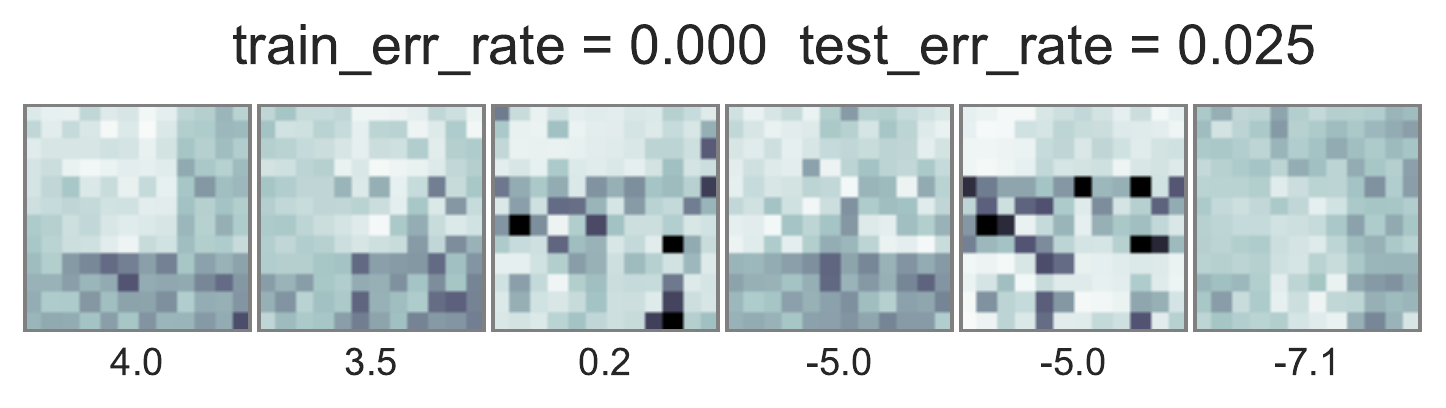}
\end{minipage}
&
\begin{minipage}{4cm}
\includegraphics[width=\textwidth]{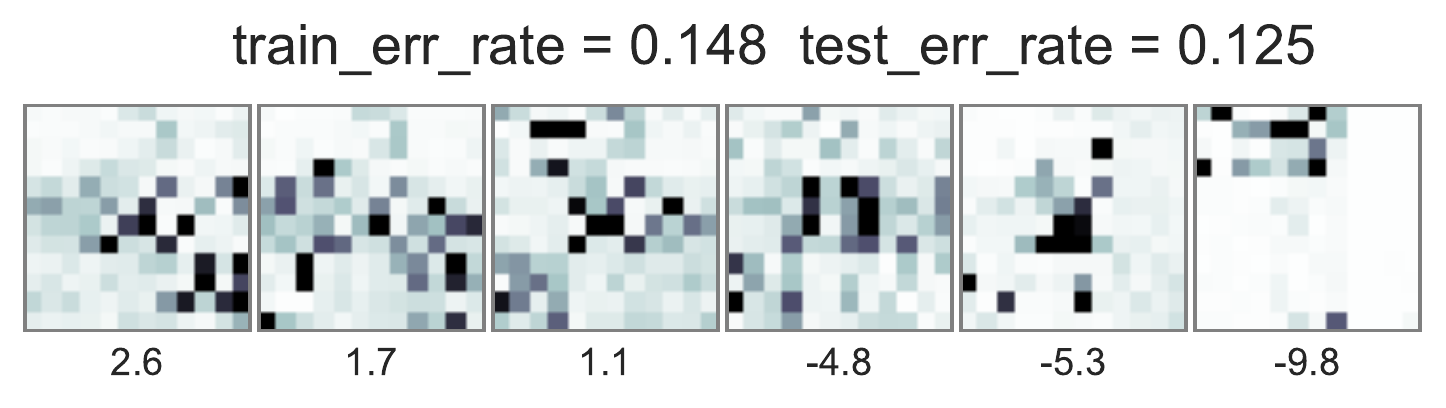}
\end{minipage}
%%%%%%%%%%%%%%%%%%%%%
\end{tabular}
}
\caption{
\textbf{Toy Bars Case Study.}
\emph{Top Row:} True topic-word parameters and example documents. Only the last 6 single bars were used to generate data $x$, but all 10 topics produce binary labels $y$ given $x$.
\emph{Remainder:}
Estimated topic-word parameters and prediction results for different training algorithms (columns).
Each row represents a setting of the penalty weights in the objective function: $w_y$ weights the supervised loss term, while $w_x$ weights the unsupervised data term.
We perform 3 separate initializations of each method with $K=6$ topics and report the one with best (lowest) training error rate.
Test set error rates are computed by using only the observed words $x_d$ and the estimated topics $\phi$ as input for each document, then computing $\pi_d$ via $f^*(x_d, \phi)$ in Eq.~\eqref{eq:def_f}.
}
\label{fig:bars_results}
\end{figure}

Two problems arise in practice with such training. First, the standard supervised LDA model sets $w_x = w_y = 1$.  However, when $x_d$ contains many words but $y_d$ has a few binary labels, the
$\log p(x)$ term dominates the objective. We see in Fig.~\ref{fig:bars_results} that the
estimated topic word parameters $\phi$ barely change between $w_x=1, w_y=0$ and $w_x=1, w_y=1$ under this standard training.

Second, the impact of observed labels $y$ on topic-word probabilities $\phi$ can be negligible.
According to the model,
when the document-topic probabilities $\pi_d$ are represented, the variables $\phi$ are conditionally independent of $y$. 
At training time the $\pi_d$ may be coerced
by direct updates using observed $y_d$ labels to make good
predictions, but such quality may \emph{not} be available at test-time, when $\pi_d$ must be updated using $\phi$ and $x_d$ alone.
Intuitively, this problem comes from the objective treating
$x_d$ and $y_d$ as ``equal'' observations when they are not. Our testing scenario always predicts labels $y_d$ from the words $x_d$. Ignoring this can lead to severe overfitting, particularly when the word weight $w_x$ is small.

\subsection{End-to-End Optimization} 
Introducing weights $w_x$ and $w_y$ can help address the first concern
(and are equivalent to providing a threshold on prediction quality).
To address the second concern, we pursue gradient-based inference of a modified version of the objective in
Eq.~\eqref{eq:def_penalized_map_objective} that respects the need to
use the same embedding of observed words $x_d$ into low-dimensional
$\pi_d$ in both training and test scenarios:
\begin{align}
\max_{\phi, \eta} 
w_y \Big( \sum_{d=1}^D \log p( y_d | f^*(x_d, \phi), \eta) \Big)
+ w_x \Big( \sum_{d=1}^D \log p( x_d | f^*(x_d, \phi), \phi) \Big)
\label{eq:def_penalized_map_objective_f}
\end{align}
The function $f^*$ maps the counts $x_d$ and topic-word parameters $\phi$ to the optimal unsupervised LDA proportions $\pi_d$. The question, of course, is how to define the function $f^*$.
% FDV: IF NEEDED, THIS IS WHERE I WOULD CUT AND SKIP TO THE
% EXPONENTIATED GRADIENT 
%
% The most costly part of the optimization is inference on the
% document-topic proportions $\pi'_{d}$ given a new document $x'_d$.
% Given a new document $x'_d$ with unknown label, we wish to first
% estimate its document topic probabilities $\pi'_d$ using the current
% topic-word parmaters $\phi$, and then predict its label $y'_d$ via
% $\sigma(\eta^T \pi'_d)$.  
One can estimate $\pi_d$ by solving a maximum a-posteriori (MAP)
optimization problem over the space of valid $K-$dimensional probability vectors
$\Delta^K$:
\begin{align}
\pi_d' = \max_{\pi_d \in \Delta^{K}} \ell(\pi_d),
\quad \ell(\pi_d) = \log p(x_d | \pi_d, \phi) + \log \mbox{Dir}(\pi_d | \alpha).
\label{eq:objective_for_pi_d}
\end{align}
%The hardness of the optimization in Eq.~\eqref{eq:objective_for_pi_d} depends on the scalar hyperparameter $\alpha$ which defines the prior $\mbox{Dir}(\pi_d | \alpha)$. Eq.~\eqref{eq:objective_for_pi_d} is non-convex if $\alpha < 1$, but convex for $\alpha \geq 1$ \citep{sontag2011complexityoflda}. However, Taddy \citep{taddy2012topicmodelmapestimation} suggests a reparameterization of the document-topic probability vector $\pi_d$ as the \emph{natural} parameter of its exponential family Dirichlet prior. This leads to convexity under \emph{any} valid setting of $\alpha$.

We can compute $\pi_d'$ via the \emph{exponentiated gradient} algorithm
\citep{kivinen1997exponentiated_gradient}, as described in \citep{sontag2011complexityoflda}.
We begin with a uniform probability vector, and iteratively reweight each entry by the exponentiated gradient until convergence using fixed stepsize $\xi
> 0$:
\begin{align}
\mbox{init:~~}
\pi^{0}_{d} \gets [\frac{1}{K} \ldots \frac{1}{K}].
\quad
\mbox{until converged:~~}
\pi^{t}_{dk} \gets \frac{p^t_{dk}}{ \sum_{j=1}^{K} p^t_{dj} },
\quad
p^t_{dk} = \pi^{t-1}_{dk} \cdot e^{ \xi \nabla \ell(\pi^{t-1}_{dk})}.
%\quad
%\pi^T_{dk} \triangleq f^{*}(x_d, \phi).
\label{eq:def_f}
\end{align}
We can view the final result after $T >> 1$ iterations, $\pi'_d
\approx \pi^T_d$, as a \emph{deterministic} function $f^*(x_d, \phi)$
of the input document $x_d$ and topic-word parameters $\phi$.

\paragraph{End-to-end training with ideal embedding.}
The procedure above does not directly lead to a way to estimate $\phi$ to maximize the objective in Eq.~\eqref{eq:def_penalized_map_objective_f}.
Recently,
\citet{chen2015bplda} developed \emph{backpropagation supervised LDA}
(BP-sLDA), which optimizes
Eq.~\eqref{eq:def_penalized_map_objective_f} under the extreme
discriminative setting $w_y = 1, w_x = 0$ by pushing gradients through
the exponentiated gradient updates above.
We can further estimate $\phi$ under any valid weights with this objective. 
We call this ``training with ideal embedding'', because the embedding is optimal under the unsupervised model.

\paragraph{End-to-end training with approximate embedding.}
Direct optimization of the ideal embedding function $f^*$, as done by
\cite{chen2015bplda}, has high implementation complexity and runtime cost. We find in practice that each document requires
dozens or even hundreds of the iterations in Eq.~\eqref{eq:def_f} to
converge reasonably. Performing such iterations at scale and
back-propagating through them is possible with careful C++
implementation but will still be the computational
bottleneck. Instead, we suggest an approximation: use a simpler
embedding function $f^{\lambda}(x_d, \phi)$ which has been trained to
approximate the ideal embedding. Initial experiments suggest
a simple multi-layer perceptron (MLP) recognition network architecture with one hidden layer of size $H \approx 50$ does reasonably well:
\begin{align}
f^{\lambda}_k(x_d, \phi) = \ts \mbox{softmax} \Big( \sum_{h=1}^H
\lambda^{\text{output}}_{hk} \sigma(\sum_{v=1}^V
\lambda^{\text{hidden}}_{hv} x_{dv} \phi_{kv}) \Big).
\end{align}
During training, we periodically pause our gradient descent over $\eta, \phi$ and update
$\lambda$ to minimize a KL-divergence loss between the approximate
embedding $f^{\lambda}$ and the ideal, expensive embedding
$f^*$.
%While some previous work used much simpler maximum-entropy approximate embeddings with poor results in a totally unsupervised setting \citep{yao2009SparseLDA}. We show that MLP recognition networks can work for supervised LDA.

\section{Case study: Toy bars data}
To understand how different training objectives impact both predictive
performance and interpretability of topic-word parameters, we consider
a version of the toy bars dataset inspired by
\cite{griffiths:2004:fst}, but changed so the optimal $\phi$ parameters are distinct for unsupervised LDA and supervised LDA objectives. Our dataset has 144 vocabulary words visualized as pixels in
a square grid in Fig.~\ref{fig:bars_results}. 
To generate the observed words $x$, we use 6 true topics: 3
horizontal bars and 3 vertical bars.
However, we generate label $y_d$ using an expanded set of 10 topics, where the extra topics are \emph{combinations} of the 6 bars.
Some combinations produce positive labels, but no single bar does. We train multiple initializations of each possible training
objective and penalty weight setting, and show the best run of each
method in Fig.~\ref{fig:bars_results}. Our conclusions are listed below:

\textbf{Standard training that instantiates $\pi$ can either ignore labels or overfit.}
Fig.~\ref{fig:bars_results}'s first column shows two problematic behaviors with the optimization objective in Eq.~\eqref{eq:def_penalized_map_objective}.
First, when $w_x = 1$, the topic-word parameters are
basically identical whether labels are ignored ($w_y = 0$) or included ($w_y = 1$). Second, when the
observed data is weighted very low ($w_x = 0.01$), we see severe
overfitting, where the learned embeddings at training time are not
reproducible at test time.

\textbf{Ideal end-to-end training can be more predictive but has expensive runtime.}  In contrast to the problems with standard training,
we see in the middle column of Fig.~\ref{fig:bars_results} that using
the ideal test-time embedding function $f^*$ also during training can produce much lower error rates on heldout
data. Varying the data weight $w_x$ interpolates between interpretable topic-word parameters $\phi$ and good predictions. One caveat to ideal embedding is its expensiveness:  Completing 100 sweeps through this 1000 document toy dataset takes about 2.5 hours using our vectorized pure Python with \texttt{autograd}.

\textbf{Approximate end-to-end training is much cheaper and often does
  as well.}  We see in the far right column of
Fig.~\ref{fig:bars_results} that using our proposed approximate
embedding $f^{\lambda}$ often yields similar predictive power and
interpretable topic-word parameters when $w_x > 0$. Furthermore, it is about 3.6X faster to train due to avoiding the expensive embedding iterations at every document.

\begin{table}
\begin{tabular}{l | rrrrr | r}
\toprule
{} & approx $f^{\lambda}$ & 
ideal $f^{*}$& 
ideal $f^{*}$&
ideal $f^{*}$&
ideal $f^{*}$& BoW 
\\
{} & $w_x=0$ & $w_x=0$ & $w_x=0.01$ & $w_x=1$ &  $w_x=1$ & 
\\
(prevalence) DRUG & $w_y=1$ & $w_y=1$ & $w_y=1$ & $w_y=1$ &  $w_y=0$ & {}
\\
\midrule
(0.215)         citalopram &                   0.65 &                 0.64 &                   0.63 &                 0.62 &                 0.61 &            0.72 \\
(0.135)         fluoxetine &                   0.66 &                 0.64 &                   0.64 &                 0.63 &                 0.63 &            0.76 \\
(0.133)         sertraline &                   0.66 &                 0.66 &                   0.63 &                 0.63 &                 0.63 &            0.75 \\
(0.119)          trazodone &                   0.64 &                 0.66 &                   0.64 &                 0.61 &                 0.62 &            0.65 \\
(0.115)          bupropion &                   0.64 &                 0.64 &                   0.59 &                 0.56 &                 0.58 &            0.71 \\
(0.070)      amitriptyline &                   0.77 &                 0.76 &                   0.77 &                 0.75 &                 0.75 &            0.78 \\
(0.059)        venlafaxine &                   0.64 &                 0.62 &                   0.62 &                 0.61 &                 0.61 &            0.73 \\
(0.059)         paroxetine &                   0.68 &                 0.73 &                   0.74 &                 0.76 &                 0.75 &            0.76 \\
(0.047)        mirtazapine &                   0.70 &                 0.69 &                   0.70 &                 0.71 &                 0.70 &            0.67 \\
(0.046)         duloxetine &                   0.71 &                 0.69 &                   0.70 &                 0.69 &                 0.70 &            0.74 \\
(0.041)       escitalopram &                   0.65 &                 0.62 &                   0.61 &                 0.61 &                 0.61 &            0.80 \\
(0.038)      nortriptyline &                   0.71 &                 0.73 &                   0.70 &                 0.70 &                 0.71 &            0.71 \\
(0.007)        fluvoxamine &                   0.70 &                 0.72 &                   0.74 &                 0.77 &                 0.76 &            0.93 \\
(0.007)         imipramine &                   0.40 &                 0.56 &                   0.50 &                 0.48 &                 0.48 &            0.82 \\
(0.006)        desipramine &                   0.47 &                 0.57 &                   0.54 &                 0.57 &                 0.54 &            0.72 \\
(0.003)         nefazodone &                   0.71 &                 0.65 &                   0.71 &                 0.72 &                 0.72 &            0.80 \\
\bottomrule
\end{tabular}
\caption{
Heldout AUC scores for prediction of 16 drugs for treating depression. Each drug decision is independent, since multiple drugs might be given to a patient. ``BoW'' is logistic regression using $x_d$ as features.
}
\label{table:psychtraj_results}
\end{table}

\section{Case study: Predicting drugs to treat depression}

We study a cohort of 875080 encounters from 49322 patients drawn from two large academic medical centers 
%and their affiliated outpatient networks
with at least one ICD9 diagnostic code for major depressive
disorder (ICD9s 296.2x or 3x or 311, or ICD10 equivalent).
Each included patient had an identified successful treatment: a prescription repeated at least 3 times in 6 months with no change.
% (to deal with overlapping meds, we considered each prescription to be active for 90 days post encounter).  

We extracted all procedures, diagnoses, labs, and meds from the EHR
(22,000 total codewords). For each encounter, we built $x_d$ by concatenating count histograms from the last three months and all prior history. 
To simplify, we reduced this to the 9621 codewords that occurred in at least 1000 distinct encounters.
The prediction goal was to identify which of 16 common anti-depressants drugs would be successful for
each patient. (Predicting all 25 primaries and 166 augments is future work).
% We limited ourselves to 16 most common of the 25 standard anti-depressants.

Table~\ref{table:psychtraj_results} compares each method's area-under-the-ROC-curve (AUC) with $K=50$ topics on a held-out set of 10\% of patients. We see that our training algorithm using the ideal embedding $f^*$ improves its predictions over a baseline unsupervised LDA model as the weight $w_x$ is driven to zero. Our approximate embedding $f^{\lambda}$ is roughly 2-6X faster, allowing a full pass through all 800K encounters in about 8 hours, yet offers competitive performance on many drug tasks except for those like desipramine or imipramine for which less than 1\% of encounters have a positive label. Unfortunately, our best sLDA model is inferior to simple bag-of-words features plus a logistic regression classifier (rightmost column ``BoW''), which we guess is due to local optima.
To remedy this, future work can explore improved data-driven initializations.

%%%%%%%%%%%%%%%%%%%%%%%%%%%%%%%%%%%%%%%%%%%% References
\newpage

{\small
\linespread{1}
\bibliography{macros_for_journal_names,references}
}

\end{document}